\documentclass{article} % For LaTeX2e
\usepackage{iclr2019_conference,times}
\usepackage{graphicx}
\usepackage{amsmath}
\usepackage{dsfont}
\usepackage{caption}
\usepackage{subcaption}

\usepackage{amsmath,amsfonts,bm}

\def\eqref#1{equation~\ref{#1}}
\def\1{\bm{1}}

\def\rva{{\mathbf{a}}}

\def\rvh{{\mathbf{h}}}

\def\ermN{{\textnormal{N}}}

\def\mC{{\bm{C}}}

\def\mN{{\bm{N}}}

\def\mX{{\bm{X}}}

\DeclareMathAlphabet{\mathsfit}{\encodingdefault}{\sfdefault}{m}{sl}
\SetMathAlphabet{\mathsfit}{bold}{\encodingdefault}{\sfdefault}{bx}{n}

\def\gG{{\mathcal{G}}}

\usepackage{hyperref}
\usepackage{url}
\usepackage{wrapfig}

\newcommand{\norm}[1]{\left\lVert#1\right\rVert}

\title{Learning Graph Neural Networks with Noisy Labels}

\iclrfinalcopy
\author{Hoang NT\thanks{https://github.com/gear/denoising-gnn} , Choong Jun Jin \& Tsuyoshi Murata \\
Department of Computer Science\\
Tokyo Institute of Technology\\
\texttt{\{hoangnt,junjin.choong\}@net.c.titech.ac.jp, murata@c.titech.ac.jp} \\
}

\begin{document}

\maketitle

\begin{abstract}

We study the robustness to symmetric label noise of GNNs 
training procedures. By combining the nonlinear neural message-passing 
models (e.g. Graph Isomorphism Networks, GraphSAGE, etc.) with loss correction methods,
we present a noise-tolerant approach for the graph classification task.
Our experiments show that test accuracy can be improved 
under the artificial symmetric noisy setting. 

\end{abstract}

\section{Introduction}

\begin{wrapfigure}{r}{0.48\textwidth}
  \vspace{-2em}
   \begin{center}
      \includegraphics[width=0.48\textwidth]{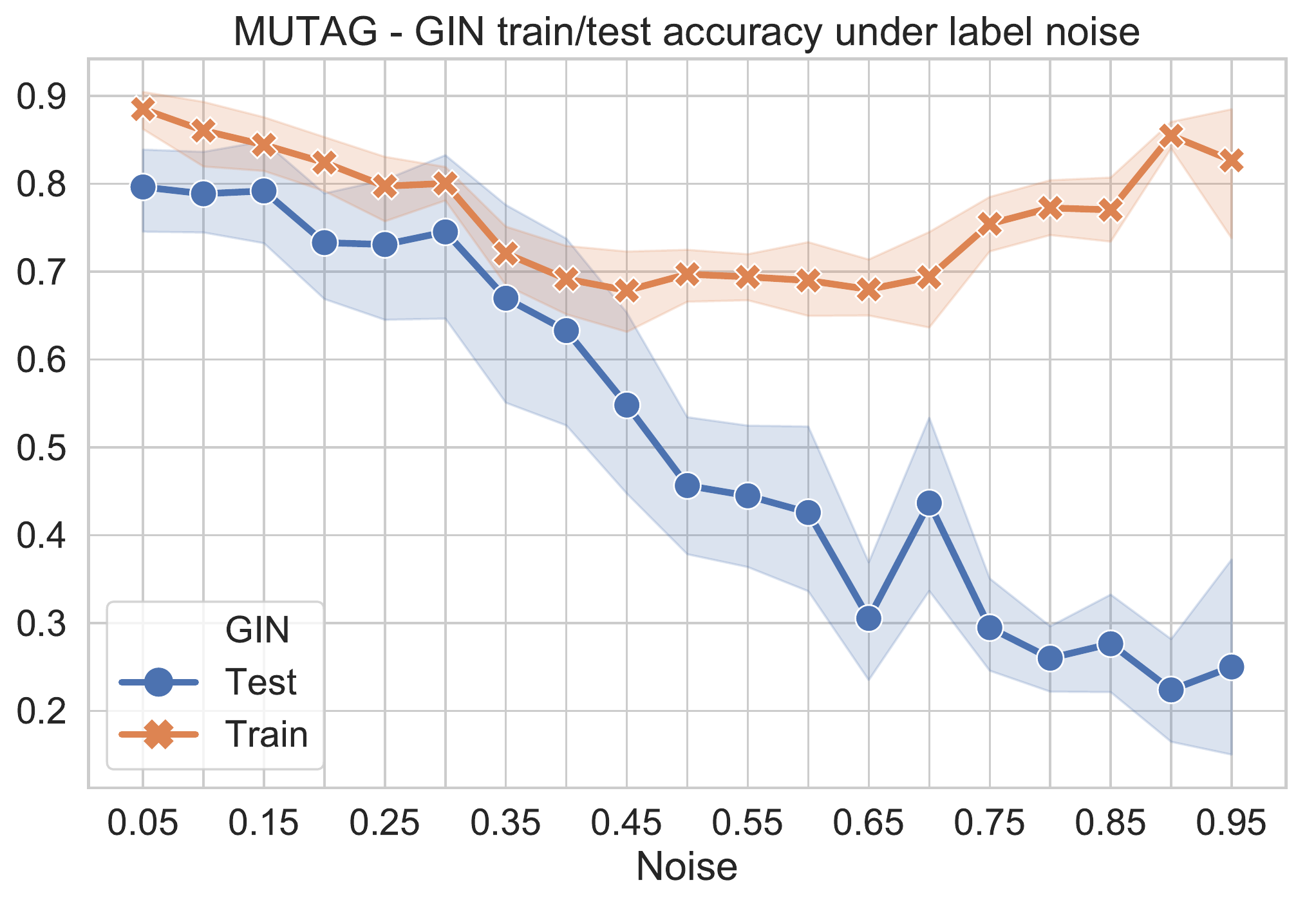}
   \end{center}
   \vspace{-1em}
  \caption{GIN model trained with increasing symmetric label noise. 
  The generalization gap increases as more noise is introduced to the training labels.}
  \label{f:mutag_gin}
  \vspace{-1em}
\end{wrapfigure}

Large datasets are beneficial to modern machine learning models, especially 
neural networks. Many studies have shown that the accuracy of machine learning
models grows log-linear to the amount of training data \citep{zhou2017brief}. 
Currently, complex machine learning models can only achieve super-human classification results when 
trained with a very large dataset. However, large datasets are usually expensive
to collect and create exact label. One solution to create large datasets is crowdsourcing, but 
this approach introduces a higher level of labeling error into the datasets as well as requires 
a lot of human resources \citep{georgakopoulos2016weakly}. As a consequence, neural networks 
are prone to very high generalization error under noisy label data.
Figure \ref{f:mutag_gin} demonstrate the accuracy results of a graph neural network
trained on MUTAG dataset. Training accuracies tend to remain high while testing
accuracies degrades as more label noise is added to the training data.

Graph neural network (GNN) is a new class of neural networks which learn from 
graph-structured data. Typically, GNNs classify graph vertices or the whole graph itself.
Given the input as the graph structure and data (e.g. feature vectors) on each 
vertex, GNNs training aim to learn a predictive model for classification. This new 
class of neural networks enables end-to-end learning from a wider range of data format.
In order to build large scale GNNs, it requires large and clean datasets. Since graph 
data is arguably harder to label than image data both at vertex-level or graph-level,
graph neural networks should have a mechanism to adapt to training label error or noise.

In this paper, we take the noise-correction approach to train a graph neural network with noisy 
labels. We study two state-of-the-art graph neural network models: \emph{Graph Isomorphism Network} \citep{xu2018how} 
and \emph{GraphSAGE} \citep{hamilton2017inductive}. Both of these models are trained under symmetric artificial label noise 
and tested on uncorrupted testing data. We then apply label noise estimation and loss 
correction techniques \citep{patrini2016loss,patrini2017making}
to propose our denoising graph neural network model (D-GNN).

\section{Method}

\subsection{Graph Neural Networks}

\paragraph{Notations and Assumption} Let $\gG = (V,E,\mX)$ be a graph with vertex set $V$, edge 
set $E$ and vertex feature vector matrix $\mX \in \mathds{R}^{\vert V \vert \times f}$, where $f$ is the 
dimensionality of vertex features. Our task is graph classification with noisy labels. Given a set of graphs: 
$\{\gG_1, \gG_2, \ldots, \gG_N\}$, their labels $\{\tilde{y}_1, \tilde{y}_2, \ldots, \tilde{y}_N\} \subset 2^m$,
we aim to learn a neural network model for graph label prediction: $y_\gG = f(\gG)$.
We assume that the training data is corrupted by a noise process $\mN$, $\ermN_{i,j}$ is 
the probability label $i$ being corrupted to label $j$. We further assume $\mN$ is symmetric, which 
corresponds to the symmetric label noise setting. Noise matrix $\mN$ is unknown, so we estimate $\mN$ 
by learning correction matrix $\mC$ from the noisy training data.

\paragraph{GNN Models} The most modern approach to the graph classification problem is to learn 
a graph-level feature vector $\rvh_\gG$. There are several ways to learn $\rvh_\gG$. \emph{GCN} approach 
by \cite{kipf2017semi} approximates the Fourier transformation of signals (feature vectors) on graphs to 
learn representations of a special vertex to use as the representative for the graph. Similar approaches 
can be founded in the context of compressive sensing. To overcome the disadvantages of GCN-like methods 
such as memory consumption and scalability, the nonlinear neural message passing method is proposed. 
\emph{GraphSAGE} \citep{hamilton2017inductive} proposes an algorithm consists of 
two operations: \texttt{aggregate} and \texttt{pooling}. \texttt{aggregate} step computes the information 
on each vertex using the local neighborhood, then \texttt{pooling} computes the output for each vertex.
These vector outputs are then used in classification at vertex-level or graph-level. More recently, 
\emph{GIN} \citep{xu2018how} model generalizes the concept in \emph{GraphSAGE} to propose a unified
message-passing framework for graph classification.

\subsection{Learning Noisy Label Data}
 
\paragraph{Surrogate Loss} Using an alternative loss function to deal with noisy label data is 
a common practice in the weakly supervised learning literature 
\citep{natarajan2013learning,biggio2012poisoning,georgakopoulos2016weakly,patrini2016loss,patrini2017making}. 
We apply the \emph{backward} loss correction procedure to graph neural network: 
$\ell^{\leftarrow} = \mC^{-1} \cdot \ell(\hat{p}(y|\gG))$. This loss can be intuitively understood 
as \emph{going backward one step} in the noise process $\mC$ \citep{patrini2017making}.

We study the \emph{symmetric} noise setting where label 
$i$ is corrupted to label $j$ with the same probability for 
$j$ to $i$ ($\ermN_{i,j} = \ermN_{j,i}$) \citep{biggio2012poisoning}. We use a
$m \times m$ symmetric Markov matrix $\mN$ to describe the noisy process with $m$ labels.
Furthermore, to simplify the experiment settings, with a given $n$ we set:
$\ermN_{i,j} = \ermN_{i,k} = n \ \forall j, k \neq i$. For example when $m=3,
n=0.2$ the noise matrix is:
$$
\mN=
  \begin{bmatrix}
    0.8 & 0.1 & 0.1 \\
    0.1 & 0.8 & 0.1 \\
    0.1 & 0.1 & 0.8
  \end{bmatrix}
$$

Matrix $N$ above can be interpreted as all labels are \emph{kept} with 
probability $0.8$ and \emph{corrupted} to other labels with probability $0.2$ 
(summation of off-diagonal elements in a row).

\subsection{Denoising Graph Neural Networks}

Formaly we define our graph neural network model as the message passing approach 
proposed by \cite{xu2018how}. The feature vector $\rvh_v$ of a vertex $V$ 
at $k$-th hop (or layer) is given by \texttt{AGGREGATE} and \texttt{COMBINE} functions:

\begin{equation}
  \begin{split}
\rva^{(k)}_{v} & = \texttt{AGGREGATE}^{(k)} (\{\rvh^{(k-1)}_u: u \in \mathcal{N}(v)\}), \\
\rvh^{(k)}_v & = \texttt{COMBINE}^{(k)} (\rvh_v^{(k-1)}, \rva_v^{(k)})
  \end{split}
\end{equation}

$\mathcal{N}(v)$ denotes the neighborhood set of vertex $v$; and $k \in [K]$ is the predefined number of 
``layers'' corresponding to network's perceptive field. The final representation 
of graph $\gG$ is calculated using a \texttt{READOUT} function. Then, we train the neural network by optimizing
the surrogate backward loss.

\begin{equation}
  \begin{split}
\rvh_{\gG} & = \texttt{READOUT} (\{\rvh^{(K)}_v: v \in \gG\}), \\
\ell^{\leftarrow}(p(y|\rvh_\gG), y_\gG) & = \mC^{-1} \cdot  \texttt{CROSS\_ENTROPY} (p(y|\rvh_\gG), y_\gG)
  \end{split}
\end{equation}

D-GNN is different from \emph{GIN} only at the surrogate loss function as 
described above. To train a D-GNN model, we first train a \emph{GIN} model 
on the noisy data for estimating $\mC$, then we train D-GNN using the estimated 
correction matrix.

We train our D-GNN model using three different noise 
estimator: Conservative (D-GNN-C), Anchors (D-GNN-A), and Exact (D-GNN-E). The 
exact loss correction is introduced for comparison purposes.
The hyperparameters of our models are set similar to GIN model in the previous
paragraph. For conservative and anchor correction matrix estimation, we train 
two models on the same noisy dataset: The first model is without loss correction
and the second model is trained using the correction matrix from the first model.
For all neural network models, we use the ReLU activation unit as the nonlinearity.

\section{Empirical Results}

We test our framework on the set of well-studied 9 datasets for the graph 
classification task: 4 bioinformatics datasets (MUTAG, PTC, NCI1, PROTEINS),
and 5 social network datasets (COLLAB, IMDB-BINARY, IMDB-MULTI, REDDIT-BINARY, REDDIT-MULTI5K)
\citep{yanardag2015deep}. We follow the preprocessing suggested by \cite{xu2018how} 
to use one-hot encoding as vertex degrees for social networks (except REDDIT datasets). 
Table \ref{t:data} gives the overview of each dataset. Since these datasets have exact 
label for each graph, we introduce symmetric label noise artificially.

\begin{wraptable}{l}{0.45\textwidth}
\vspace{-1em}
\caption{Data overview}
\label{t:data}
\resizebox{0.45\textwidth}{!}{%
\begin{tabular}{l|cccc}
\bf Dataset & \#graphs & \#classes & \#vertices \\
\hline \\
IMDB-B   & 1000 & 2 & 19.8  \\   
IMDB-M   & 1500 & 3 & 13.0  \\ 
RDT-B    & 2000 & 2 & 429.6 \\    
RDT-M5K  & 5000 & 5 & 508.5 \\       
COLLAB   & 5000 & 3 & 74.5  \\     
MUTAG    & 188  & 2 & 17.9  \\    
PROTEINS & 1113 & 2 & 39.1  \\       
PTC      & 344  & 2 & 25.5  \\  
NCI1     & 4110 & 2 & 29.8  \\   
\end{tabular}}
\vspace{-2em}
\end{wraptable}

\subsection{Noise Estimation}

\paragraph{Conservative Estimation} We estimate the corruption probability by
the Conservative Estimator described in the previous sections. For each noise 
configuration, we train the original neural network (GIN) on the
noisy data and use the neural response to fill each row of the correction 
matrix $\mC$. Table \ref{t:est_exact} gives an overview of how well the 
conservative estimation matrix diverges from the correct noise matrix.
The matrix norm $\norm{\mC-\mN}$ is the $p$-norm with $p=1$.

\paragraph{Anchor Estimation} We follow the noise estimation method introduced
in \cite{patrini2017making} (Equations (12,13)) to estimate the noise probability 
using an unseen set of samples. These anchor samples are assumed to have the correct
labels, hence they can be used to estimate the noise matrix according to 
the expressivity assumption. In our experiments, these samples are taken from
the testing data (one per class). Table \ref{t:est_exact} demonstrates the
similarity results.

\begin{table}[h]
\caption{Norm distance between conservative correction matrix estimation
$\mC^{\text{c}}$ and $\mC^{\text{a}}$ compared with true noise matrix $\mN$ when $n=0.2$}
\begin{center}
\resizebox{0.75\textwidth}{!}{%
\begin{tabular}{l|ccccc}
\bf{Dataset} (\#classes) & diag($N$) & Avg. diag($\mC^{\text{c}}$) & $\norm{\mC^{\text{c}}-N}$ & Avg. diag($\mC^{\text{a}}$) & $\norm{\mC^{\text{a}}-N}$ \\
\hline \\
IMDB-B   (2) & 0.8  & 0.99   & 0.76  & 0.77 & 0.12\\
IMDB-M   (3) & 0.8  & 0.99   & 1.14  & 0.85 & 0.30\\
RDT-B    (2) & 0.8  & 0.99   & 0.76  & 0.75 & 0.20\\
RDT-M5K  (5) & 0.8  & 0.99   & 1.90  & 0.81 & 0.10\\
COLLAB   (3) & 0.8  & 0.99   & 1.14  & 0.75 & 0.30\\
MUTAG    (2) & 0.8  & 0.99   & 0.76  & 0.74 & 0.24\\
PROTEINS (2) & 0.8  & 0.99   & 0.76  & 0.78 & 0.08\\
PTC      (2) & 0.8  & 0.99   & 0.76  & 0.63 & 0.68\\
NCI1     (2) & 0.8  & 0.99   & 0.76  & 0.74 & 0.24\\
\end{tabular}}
\end{center}
\label{t:est_exact}
\vspace{-2em}
\end{table}

\paragraph{Exact Assumption} In this experiment setting, we assume that the 
noise matrix is exactly known from some other estimation process. In practice,
such an assumption might not be realistic. However, under the symmetric noise assumption,
the diagonal of the correction matrix $\mC$ can be tuned as a hyperparameter.

\subsection{Graph Classification}

We compare our model with the original Graph Isomorphism Network (GIN) \citep{xu2018how}.
The hyperparameters are fixed across all datasets as follow: \texttt{epochs=20}, 
\texttt{num\_layers=5}, \texttt{num\_mlp\_layers=2}, \texttt{batch\_size=64}.
We keep these hyperparameters fixed for all datasets since the similar 
trend of accuracy degradation is observed independently of hyperparameter tuning. 
Besides GIN, we consider GraphSAGE model \citep{hamilton2017inductive} under the same noisy 
setting. We use the default setting for GraphSAGE as suggested in the original paper.

\begin{table*}[h]
  \centering
  \caption{Classification results at symmetric noise, when $n=0.2$ (80\% data has correct labels).
  We calculate the mean and std of accuracy score on test data for 10 runs each configuration. Bold font indicates
  improvement compared to the original model.}
  \label{tab:1}
  \begin{center}
  \resizebox{\textwidth}{!}{%
  \begin{tabular}{lccccccccccc}
                                 & MUTAG   & IMDB-M    & RDT-B   & RDT-M5K & COLLAB    & IMDB-B   & PROTEINS    & PTC         & NCI1 \\
    \hline \\
    GIN                          & .7327   & .4476     & .6695   & .3677   & .6544     & .6573    & .6257       & .4824       & .6472 \\
    GraphSAGE                    & .7072   & .4373     &   -     & -       & -         & .6410    & .6583       & .4892       & .6053 
    \vspace{0.5em}\\
    \hline \vspace{-0.5em} \\
    D-GNN-C                      &  .5727  &  \bf.4747 & .5005   & .2000   &  .5979    & \bf.6940 & \bf.6693    & \bf.5557       &  .6170 \\
    D-GNN-A                      &  .7102  &  \bf.4505 & .5307   & .2000   &  \bf.6917 & \bf.7088 & \bf.6769    & \bf.5001       & .6405  \\
    D-GNN-E                      &  .7002  &  \bf.4633 & .5270   & .2022   &  \bf.6960 & \bf.7190 & \bf.6917    & \bf.5235       & \bf.6638 \\
  \end{tabular}}
\end{center}
\vspace{-1em}
\end{table*}

\begin{figure}[h]
\centering
\begin{subfigure}{0.33\textwidth}
  \centering
  \includegraphics[width=\textwidth]{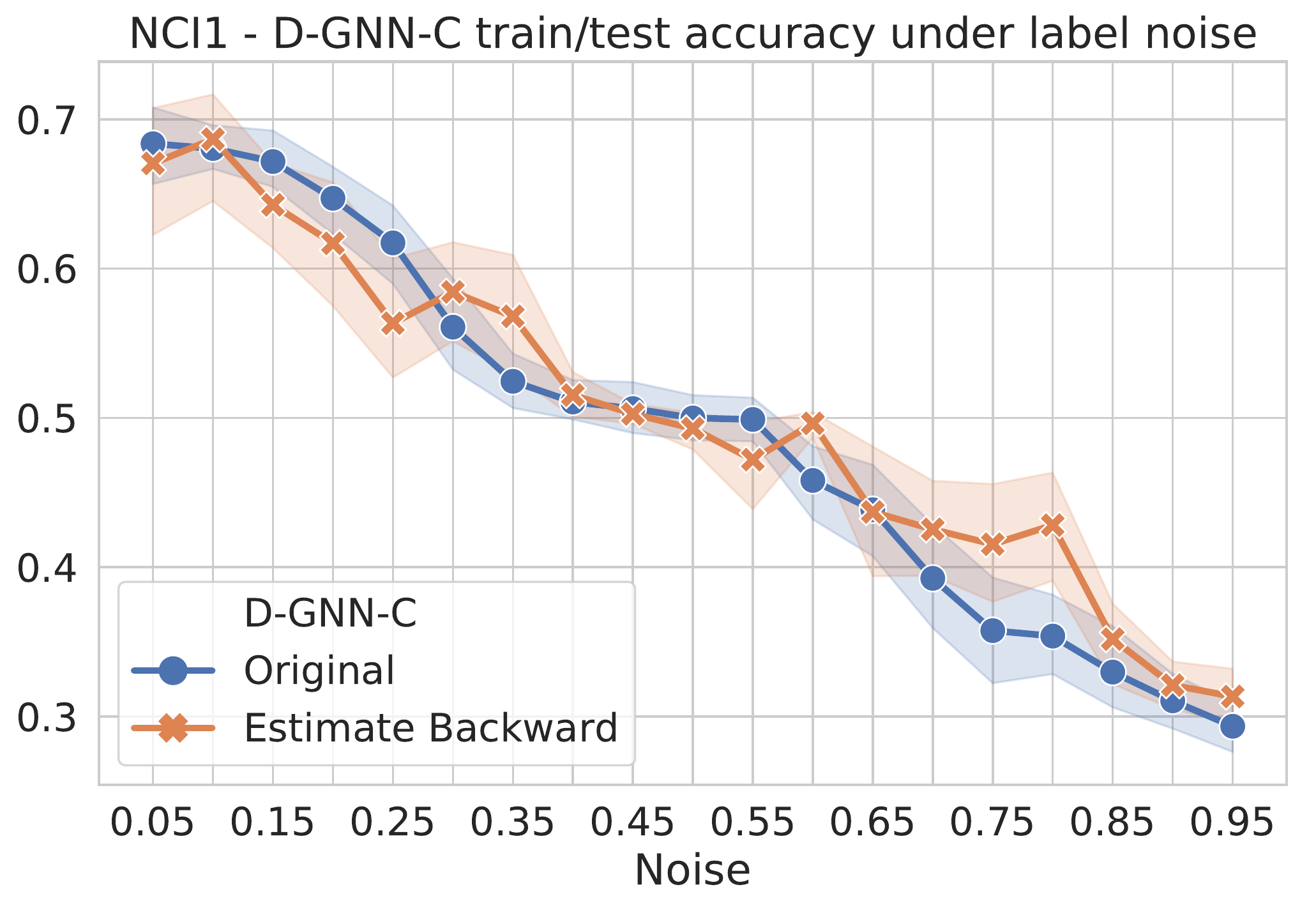}
  \caption{NCI1}
  \label{f:nci1_est_back}
\end{subfigure}%
\begin{subfigure}{0.33\textwidth}
  \centering
  \includegraphics[width=\textwidth]{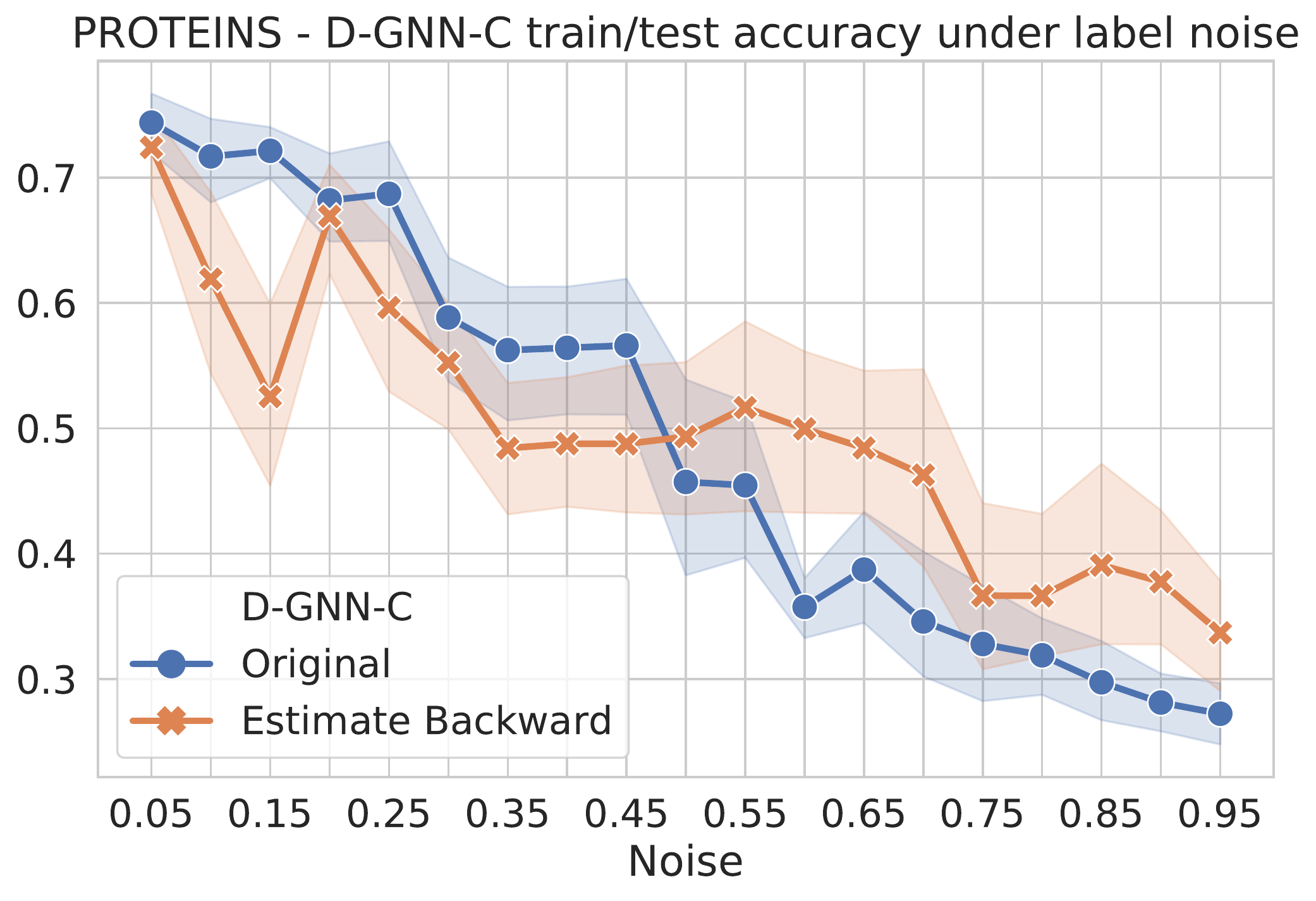}
  \caption{PROTEINS}
  \label{f:proteins_est_back}
\end{subfigure}
\begin{subfigure}{0.33\textwidth}
  \centering
  \includegraphics[width=\textwidth]{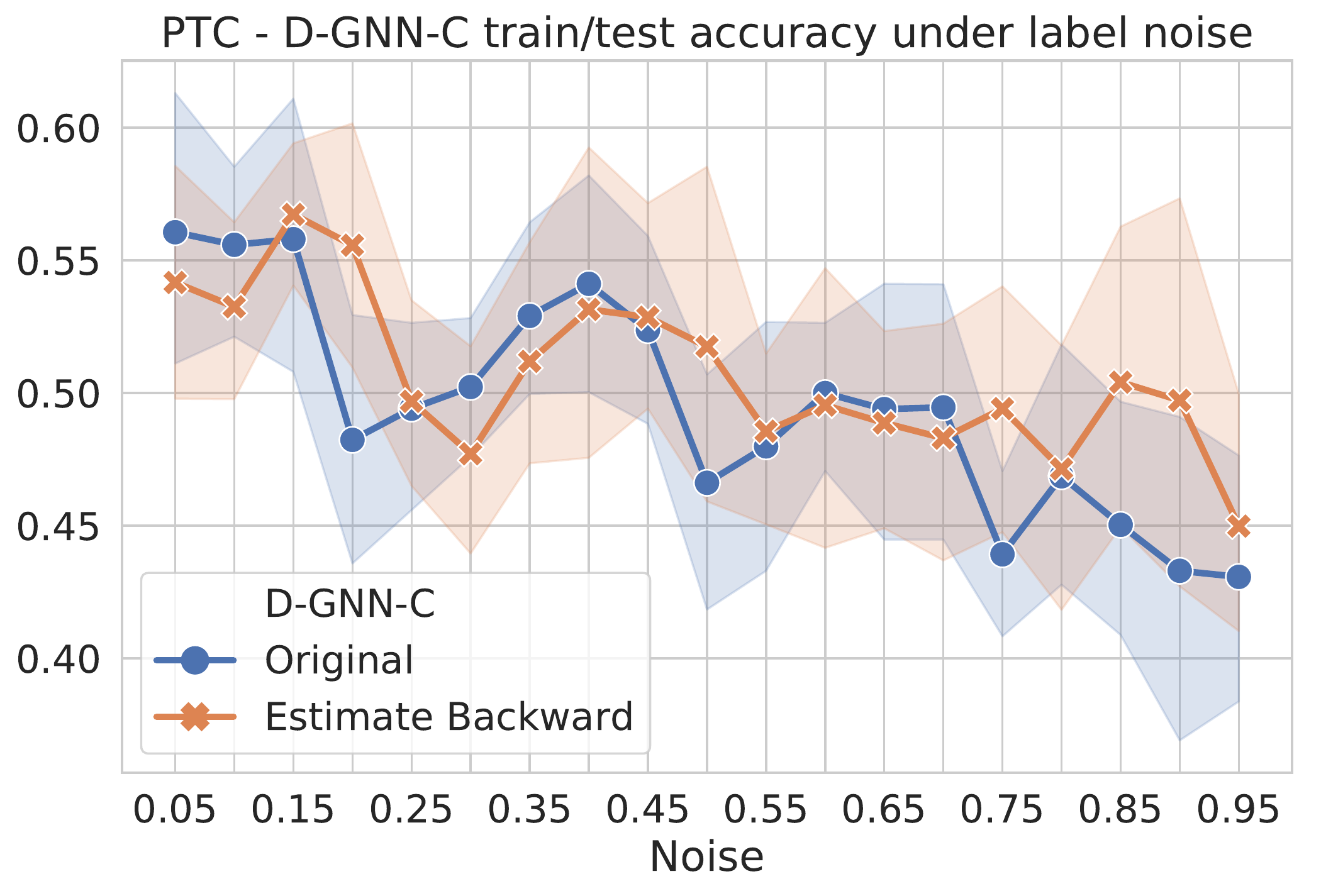}
  \caption{PTC}
  \label{f:ptc_est_back}
\end{subfigure}
\caption{Denoising results on bioinformatics datasets. X-axis presents the test accuracies.}
\label{f:denoising_est_back}
\end{figure}

We fix the noise rate at 20\% for the experiments in Table \ref{tab:1} and report 
the mean accuracy after 10 fold cross validation run.  
The worst performance variance of our model is the conservative estimation model. Due to the 
overestimation of softmax unit within the cross-entropy loss, the model's confidence 
to all training data is close to $1.0$. Such overconfidence leads to wrong correction
matrix estimation, which in turn leads to worse performance (Table \ref{t:est_exact}).
In contrast to D-GNN-C, D-GNN-A and D-GNN-E have consistently outperformed the original
model. Such improvement comes from the fact that the correction matrix $\mC$ is correctly 
approximated. Figure \ref{f:denoising_est_back} suggests that the D-GNN-C model might
work well under the higher label noise settings. 

\section{Conclusion}

In this paper, we have introduced the use of loss correction for Graph Neural Networks 
to deal with symmetric graph label noise. We experimented on 
two different practical noise estimatation methods and compare them to the case when we know 
the exact noise matrix. Our empirical results show some improvement 
on noise tolerant when the correction matrix $\mC$ is correctly estimated. 
In practice, we can consider $\mC$ as a hyperparameter and tune it following some clean validation 
data. 

\subsubsection*{Acknowledgments}

This work was supported by JSPS Grant-in-Aid for Scientific Research (B) (Grant Number 17H01785) 
and JST CREST (Grant Number JPMJCR1687).

\bibliography{iclr2019_conference}

\begin{thebibliography}{10}
\providecommand{\natexlab}[1]{#1}
\providecommand{\url}[1]{\texttt{#1}}
\expandafter\ifx\csname urlstyle\endcsname\relax
  \providecommand{\doi}[1]{doi: #1}\else
  \providecommand{\doi}{doi: \begingroup \urlstyle{rm}\Url}\fi

\bibitem[Biggio et~al.(2012)Biggio, Nelson, and Laskov]{biggio2012poisoning}
Battista Biggio, Blaine Nelson, and Pavel Laskov.
\newblock Poisoning attacks against support vector machines.
\newblock \emph{arXiv preprint arXiv:1206.6389}, 2012.

\bibitem[Georgakopoulos et~al.(2016)Georgakopoulos, Iakovidis, Vasilakakis,
  Plagianakos, and Koulaouzidis]{georgakopoulos2016weakly}
Spiros~V Georgakopoulos, Dimitris~K Iakovidis, Michael Vasilakakis, Vassilis~P
  Plagianakos, and Anastasios Koulaouzidis.
\newblock Weakly-supervised convolutional learning for detection of
  inflammatory gastrointestinal lesions.
\newblock In \emph{2016 IEEE international conference on imaging systems and
  techniques (IST)}, pp.\  510--514. IEEE, 2016.

\bibitem[Hamilton et~al.(2017)Hamilton, Ying, and
  Leskovec]{hamilton2017inductive}
Will Hamilton, Zhitao Ying, and Jure Leskovec.
\newblock Inductive representation learning on large graphs.
\newblock In \emph{Advances in Neural Information Processing Systems}, pp.\
  1024--1034, 2017.

\bibitem[Kipf \& Welling(2017)Kipf and Welling]{kipf2017semi}
Thomas~N. Kipf and Max Welling.
\newblock Semi-supervised classification with graph convolutional networks.
\newblock In \emph{International Conference on Learning Representations
  (ICLR)}, 2017.

\bibitem[Natarajan et~al.(2013)Natarajan, Dhillon, Ravikumar, and
  Tewari]{natarajan2013learning}
Nagarajan Natarajan, Inderjit~S Dhillon, Pradeep~K Ravikumar, and Ambuj Tewari.
\newblock Learning with noisy labels.
\newblock In \emph{Advances in neural information processing systems}, pp.\
  1196--1204, 2013.

\bibitem[Patrini et~al.(2016)Patrini, Nielsen, Nock, and
  Carioni]{patrini2016loss}
Giorgio Patrini, Frank Nielsen, Richard Nock, and Marcello Carioni.
\newblock Loss factorization, weakly supervised learning and label noise
  robustness.
\newblock In \emph{International conference on machine learning}, pp.\
  708--717, 2016.

\bibitem[Patrini et~al.(2017)Patrini, Rozza, Krishna~Menon, Nock, and
  Qu]{patrini2017making}
Giorgio Patrini, Alessandro Rozza, Aditya Krishna~Menon, Richard Nock, and
  Lizhen Qu.
\newblock Making deep neural networks robust to label noise: A loss correction
  approach.
\newblock In \emph{Proceedings of the IEEE Conference on Computer Vision and
  Pattern Recognition}, pp.\  1944--1952, 2017.

\bibitem[Xu et~al.(2019)Xu, Hu, Leskovec, and Jegelka]{xu2018how}
Keyulu Xu, Weihua Hu, Jure Leskovec, and Stefanie Jegelka.
\newblock How powerful are graph neural networks?
\newblock In \emph{International Conference on Learning Representations}, 2019.
\newblock URL \url{https://openreview.net/forum?id=ryGs6iA5Km}.

\bibitem[Yanardag \& Vishwanathan(2015)Yanardag and
  Vishwanathan]{yanardag2015deep}
Pinar Yanardag and SVN Vishwanathan.
\newblock Deep graph kernels.
\newblock In \emph{Proceedings of the 21th ACM SIGKDD International Conference
  on Knowledge Discovery and Data Mining}, pp.\  1365--1374. ACM, 2015.

\bibitem[Zhou(2017)]{zhou2017brief}
Zhi-Hua Zhou.
\newblock A brief introduction to weakly supervised learning.
\newblock \emph{National Science Review}, 5\penalty0 (1):\penalty0 44--53,
  2017.

\end{thebibliography}
\bibliographystyle{iclr2019_conference}

\end{document}